\documentclass[sigconf]{aamas} 
\renewcommand\footnotetextcopyrightpermission[1]{} 
\usepackage{balance} 
\usepackage{algpseudocode}
\usepackage{algorithm}
\usepackage{natbib}

\title {Evaluating the Robustness of Off-Road Autonomous Driving Segmentation against Adversarial Attacks: A Dataset-Centric analysis}

    \author{Pankaj Deoli}
\affiliation{
  \institution{Robotics Research Lab, RPTU}
  \city{Kaiserslautern}
  \country{Germany}}
\email{deoli@cs.uni-kl.de}

\author{Rohit Kumar}
\affiliation{
  \institution{Robotics Research Lab, RPTU}
  \city{Kaiserslautern}
  \country{Germany}}
\email{r_kumar22@cs.uni-kl.de}

\author{Axel Vierling}
\affiliation{
  \institution{Robotics Research Lab, RPTU}
  \city{Kaiserslautern}
  \country{Germany}}
\email{vierling@cs.uni-kl.de}

\author{Karsten Berns}
\affiliation{
  \institution{Robotics Research Lab, RPTU}
  \city{Kaiserslautern}
  \country{Germany}}
\email{berns@cs.uni-kl.de}

\begin{abstract}
This study investigates the vulnerability of semantic segmentation models to adversarial input perturbations, in the domain of off-road autonomous driving. Despite good performance in generic conditions, the state-of-the-art classifiers are often susceptible to (even) small perturbations, ultimately resulting in inaccurate predictions with high confidence. Prior research has directed their focus on making models more robust by modifying the architecture and training with noisy input images, but has not explored the influence of datasets in adversarial attacks. Our study aims to address this gap by examining the impact of non-robust features in off-road datasets and comparing the effects of adversarial attacks on different segmentation network architectures. To enable this, a robust dataset is created consisting of only robust features and training the networks on this robustified dataset. We present both qualitative and quantitative analysis of our findings, which have important implications on improving the robustness of machine learning models in off-road autonomous driving applications. Additionally, this work contributes to the safe navigation of autonomous robot Unimog U5023 in rough off-road unstructured environments by evaluating the robustness of segmentation outputs. The code is publicly available at \textit{\textbf{\url{https://github.com/rohtkumar/adversarial_attacks_on_segmentation}}}
\end{abstract}


\begin{document}


\pagestyle{fancy}
\fancyhead{}


\maketitle 


\section{Introduction}

Technology has come a long way since the development of the first automobile by Karl Benz and Gottlieb Daimler. With the advent of high-stakes applications, such as autonomous driving, which heavily relies on Deep Neural Networks (DNNs) for perception, people are increasingly fascinated by the capabilities of machines. Today, DNNs form the backbone of all computer vision solutions, thereby providing the industry with human-like capabilities in the field of computer vision. However, safety concerns have arisen in relation to autonomous driving with respect to DNNs, particularly due to the vulnerability of such systems to adversarial attacks. Adversarial attacks refer to intentionally created inputs that trick well-trained DNNs, especially when the adversary needs to make only minor changes to an input that has already been successfully processed. On a broad level, these attacks can be categorized along 3 primary axes i.e. their effect or influence on the classifier, any safety violations and individual specificity. These can be further sub-categorized as \textit{White box} or \textit{Black box} attacks, depending on whether the attacker has access to the model's parameters. Moreover, these attacks typically involve carefully constructed adversarial examples at the level of pixels, which manipulate the physical environment and thus result in physical consequences. \\  
Adversarial examples represent a major area of research interest due to both safety concerns and the need for explainability. Significant research and developmental efforts have been carried out in the domain of Adversarial examples for general classification tasks but not so much for off-road autonomous driving. Further, these improvements have been oriented towards urban contexts since addressing man-made structural entities is relatively simple. On the other hand, great effort has been put towards segmenting paths, trees, and plants in off-road areas, but little towards enhancing the robustness of autonomous driving systems to adversarial examples in such environments.\\
This work addresses this gap by evaluating the effects of these adversarial attacks for semantic segmentation in the context of off-road autonomous driving. The main contributions of the paper are summarized as follows:
\begin{itemize}
    \item An off-road robust dataset has been created (by adapting \cite{ilyas2019adversarial}) consisting of only robust features.
    \item Analysis of this robustified dataset on two (established) State-of-the-art (SOTA) semantic segmentation networks i.e. U-Net and Link-Net.
    \item Networks comparison and analysis on standard dataset vs Robustified dataset
\end{itemize}

The workflow is as follows: section II provides related works. The goal of section III is to present the methodology, the datasets used followed by architectural, implementation and training details. Section IV provides the experimentation details along-with quantitative and qualitative comparison whereas the evaluation of this work is done in section V. The strategic transfer of this work to the Unimog U5023 is mentioned in section VI, whereas, the work is concluded with future scope in section VII.


\section{Related Works}

The work done in the field of adversarial examples suggests a variety of explanations ranging from theoretical models by \cite{schmidt2018adversarially}, \cite{bubeck2018adversarial} to arguments based on the concentration of measure in high dimensions \cite{gilmer2018adversarial}, \cite{mahloujifar2018curse}. An extensive adversarial robustness evaluation of real-time semantic segmentation models is presented by \cite{rossolini2022realworld} wherein, different attacks are evaluated. On the contrary, certified defences against adversarial patch attacks was presented by \cite{yatsura2023certified} with the introduction of demasked smoothing which can be applied to any semantic segmentation model. Using multi-sensor setup, the authors of \cite{lee_2021} proposed an adversarially robust fusion method by developing a random feature fusion strategy for preserving multi-sensor fusion features. However, the aforementioned researches frequently falls short of accurately capturing the behaviours we observe in practise. The works done focus on carefully crafting adversarial examples and then adding it to the image but do not take into consideration the non-robust features (inherently present in an image) which might affect the overall performance. According to an earlier research on the subject, adversarial examples are typically seen as aberrations caused by the large dimensions of input space or statistical irregularities in the training data \cite{mahloujifar2018curse}, \cite{szegedy2014intriguing}, \cite{goodfellow2015explaining}. In the context of Autonomous Driving (AD), many ideas for understanding adversarial examples have been put forth, ranging from high-dimensional statistical phenomenon to finite-sample overfitting \cite{gilmer2018adversarial}, \cite{fawzi2018adversarial} and \cite{tanay2016boundary}. Hence, the robustness of the adversarial approach can be seen as a target that can be separated from maximizing the accuracy to be pursued either through enhanced regularization techniques or pre/post processing of network inputs/outputs \cite{uesato2018adversarial}, \cite{carlini2017adversarial}, \cite{he2017adversarial}.

\section{Approach and implementation}
The following section gives an overview on the methodology and approach taken. It begins with an introduction of the types of adversarial attacks, followed by dataset exploration, baseline model selection criteria, and network training. 
\subsection{Gradient-based adversarial attacks}
Gradient-based adversarial attacks exploit the use of gradients in ML models to generate examples. These gradients have been utilized in different manners for obtaining the perturbation vector, such that the new adversarial example has a greater tendency towards being misclassified. \textit{Fast Gradient Sign Method (FGSM)} \cite{huang2017adversarial} is one of the simplest and most effective adversarial attacks wherein the gradients of the loss functions are calculated (w.r.t the input data) and then perturbating the input data in the direction of the sign of the gradient.

\begin{equation}
    W^T x^{adv} + b = W^Tx + W^T \delta + b    
\label{eq_1}
\end{equation}

where, $x$ denotes normal examples without adding the perturbations, $\delta$ is referred to as perturbation in adversarial examples and indicates the adversarial perturbations added to normal examples. The adversarial examples are fed into the ML model and $x$ multiplies the parameter matrix $W^T$ given as equation \ref{eq_1}. \\
\textit{Basic Iterative Method (BIM)} \cite{kurakin2017adversarial}, an iterative version of FGSM, involves applying the FGSM multiple times with a small perturbation in each iteration (which can increase the likelihood of success). A stronger version of BIM is \textit{Projected Gradient Descent (PGD)} \cite{kurakin2017adversarial} which includes a projection step to ensure that the perturbed input data remains within a certain range of values thereby making the attack more effective against well-trained models. The basic PGD algorithm simply iterates the updates:
\begin{equation}
    Repeat: 
    \delta:=\mathcal{P}\left(\delta+\alpha \nabla_\delta \mathcal{L}\left(h_\theta(x+\delta), y\right)\right)
\end{equation}

where $\mathcal{P}$ denotes the projection onto the ball of interest, $h_\theta$ is a neural network, $\delta$ denotes the lower bound or delta, $\theta$ denotes the model parameters and $\alpha$, the step size.  Furthermore, the following loss function is minimized during  training, where $\Delta$ is a set of perturbations that we want our model to be invariant to, such as the $L^2$ and $L^\infty$ perturbations.
\begin{equation}
    \min _\theta \max _{\delta \in \Delta} \mathcal{L}(x+\delta, y ; \theta) 
\end{equation} 

And therefore due to this efficiency, we chose \textbf{PGD attack} (with $L^{2}$ and $L^{\infty}$ bounds) for our task at hand. 

\subsection{Baseline model selection}
Before deciding on the appropriate model, various network architectures were studied. These included FCN16 \cite{long2015fully}, UNet \cite{ronneberger2015unet}, DeepLab \cite{chen2017deeplab}, PNPNet \cite{liang2020pnpnet} and LinkNet \cite{Chaurasia_2017}. Comparing all, UNet and PNPNet employs novel loss weighting algorithm that places a higher weight near the edges of segmented objects. This weighting scheme helps the model to learn pixel-to-pixel mapping and to segment cells in the image in a discontinuous fashion and makes them identifiable. U-Net performs image localization by predicting the image pixel by pixel and
works with fewer training images to produce precise segmentation results. As the network
works with a smaller dataset and still produces comparable results, the network was
chosen for this study. Unlike many DNNs used for segmentation, Link-Net
links each encoder with a decoder such that the input of each encoder layer is also bypassed to the output of its corresponding decoder so as to recover lost spatial information that can be used by the decoder and its up-sampling operations. Since decoder is sharing knowledge learned by encoder at every layer, the decoder can use fewer parameters. Based on the limitations of our small dataset and the observed performance of various networks with small datasets, \textbf{U-Net} and \textbf{LinkNet} were selected as the baseline models.

\subsection{Datasets}
A good dataset provides the foundation for the model to learn and make accurate predictions and thereby should contain a wide variety of data (representative) that reflects an entire range of variability in the problem domain. When compared to widely used urban datasets for autonomous driving such as KITTI (12919 images) \cite{geiger_kitti}, Cityscapes \cite{cordts2016cityscapes} (2500 images), Waymo \cite{sun2020scalability} (1000 images), off-road datasets lack distinctiveness and quantity. Prominent challenges when working with off-road environments include occlusion, changing light conditions, and objects with in-definitive shapes and sizes. Freiburg forest \cite{valada_abhinav_2017} (366 pixel-wise annotated images) dataset is widely used for semantic scene understanding of off-road environments with 6 classes. The Yamaha CMU Off-road dataset (YCOR) \cite{YCOR_dataset}, on the other hand, contains (1076 images with 7 classes) collected in 4 different seasons.

\begin{table}[htp!]
\centering
\caption{Combined Off-road dataset classes}
\begin{tabular}{|l|l|} 
\hline
\textbf{Class}          & \textbf{Label} \\ 
\hline
Background        & 0 \\
Vegetation        & 1 \\
Traversable grass & 2\\
Smooth trail  & 3 \\
Obstacle & 4\\
Sky & 5\\
Rough trail & 6 \\
Puddle & 7\\
 \begin{tabular}[c]{@{}l@{}}Non Traversable\\Vegetation\end{tabular} &8\\
Tree & 9\\

\hline
\end{tabular}
\label{tab:offroad-table}
\end{table}
To increase diversity, sample size, and model transferability, datasets can be carefully combined  to improve the ML model's generality. But this should be done carefully so as to negate the effects of introducing biases and inconsistencies. For our study, the above-mentioned two off-road datasets were carefully combined (a total of 1442 images) with the final 10 classes as seen from table \ref{tab:offroad-table}. For model training, the train/val/test distribution was selected as 794/360/288 images respectively.  

\subsection{Network training}
After deciding the networks, adversarial attacks and datasets, the models were trained with resnet50 backbone. For this, 4 Nvidia GeForce RTX 2080 Ti GPU (total capacity of 44 GB) were used with an Intel(R) Xeon (R) Gold 6126 CPU @2.60 GHz. For the task of adversarial attacks, the segmentation model module in Tensorflow is used in order to understand the performance of the training both quantitatively and qualitatively. The experiments however, were divided in 4 stages i.e.  
\begin{itemize}
    \item \textbf{Standard training on the merged dataset} - This involves training the models on the merged dataset using standard training configurations, without any adversarial attacks. This allows us to assess the models performance on an unbiased environment. 
    \item \textbf{Adversarial training (\textit{PGD} ($L^2$ and $L^\infty$))} - wherein the networks are trained with adversarial attacks with the adversarial training configurations.
    \item \textbf{Robustifying the merged dataset} - i.e. dividing the dataset according to robust and non-robust features and training the models only on the robust features. The goal here is to remove any noise or irrelevant features that may affect the models performance.
    \item \textbf{Standard training on Robustified dataset} -  wherein the standard training is carried out on the robustified dataset with the goal to assess the models performance when trained on a dataset that has been modified to remove non-robust features.  
\end{itemize}

The configurations for training, testing, adversarial attacks can be seen from table \ref{table_1}. 
\begin{table}[htbp!]
\centering
\small
\caption{Hyperparameter setup for different experiments}
\begin{tabular}{|l|l|}
\hline
                            & \textbf{Unet/Linknet}                                                                                                                                                                                                                        \\ \hline
\textbf{Training Config.}             & \begin{tabular}[c]{@{}l@{}}\textbf{Optimizer} - Adam \\ \textbf{Learning Rate} - 1$\times$$e^{-4}$\\ \textbf{Loss} -  CategoricalCrossEntropy\\ \textbf{Metrics} - IoU, Accuracy\\ \textbf{Augmentation} -  Random flip, \\ Random hue, Contrast, Brightness\\ \textbf{Epochs} - 100\end{tabular} \\ \hline
\textbf{Test Config.}                 & \begin{tabular}[c]{@{}l@{}}\textbf{Loss} - CategoricalCrossEntropy\\ \textbf{Metrics} - IoU, Accuracy\\ \end{tabular}                                                                                                            \\ \hline
\textbf{Adversarial Training Config.}        & \begin{tabular}[c]{@{}l@{}}\textbf{Optimizer} - Adam\\ \textbf{Learning Rate} - 1$\times$$e^{-4}$\\ \textbf{Loss} - CategoricalCrossEntropy\\ \textbf{Metrics} - IoU, Accuracy\\ \textbf{Attack} - PGD $L^2$, PGD $L^\infty$\\ \textbf{epsilon} - 10\\ \textbf{alpha} - 0.1\end{tabular}                           \\ \hline
\textbf{Robustifier/Non-Robustifier} & \begin{tabular}[c]{@{}l@{}}\textbf{Attack} - PGD $L^2$, PGD $L^\infty$,\\ \textbf{alpha} - 0.1\end{tabular}                                                                                                                                                \\ \hline
\end{tabular}
\label{table_1}
\end{table}

\section{Experimentation}

\subsection{Standard training on the merged dataset} \label{standard-test}
Before starting the subsequent studies, it is important to perform experimentation for baseline results comparison. Having no comparison measure (with merged dataset), this helps us to derive the baseline with which we can compare the later results. The models (both UNet and LinkNet) are trained for 100 epochs with the configurations mentioned in table \ref{table_1}. Table \ref{tab:table_2} shows the quantitative results, wherein, UNet achieves an IoU of 73\% during training and 69\% during validation. LinkNet achieves an IoU of 71\% and 69\% during training and validation respectively. For baselining of the results, these trained models (UNet and LinkNet) are then evaluated against \textit{PGD $L^2$} and \textit{PGD $L^\infty$} attacks on the test dataset. The results are presented in table \ref{tab:table_2} wherein it can be observed that the performance of both the networks went considerably down. The networks fails to learn the non adversarial features and therefore predicts the wrong output.\\ Having baseline results for comparison we now train the networks with the aforementioned attacks, the details of which are mentioned in the next section. 

\begin{table}[htp!]
\centering
\caption{Model is trained in standard setup for 100 epochs. The loss represents the average CategoricalCross entropy loss after 100 epochs and IoU represents the average after 100 epochs.}
\begin{tabular}{|l|l|l|}
\hline
 & \textbf{UNet} & \textbf{LinkNet} \\ \hline
\textbf{Train IoU}   & \textbf{0.73} & \textbf{0.71}             \\ \hline
\textbf{Train Loss}  & 0.16 & 0.17          \\ \hline
\textbf{Val IoU}     & 0.69 & 0.69           \\ \hline
\textbf{Val Loss}    & 0.32 & 0.37        \\ \hline
\textbf{Test IoU (PGD $L^2$)} & 0.48 & 0.60               \\ \hline
\textbf{Test IoU (PGD $L^\infty$)} & 0.39 & 0.58               \\ \hline
\textbf{Test Loss (PGD $L^2$)}   & 0.59 & 0.76         \\ \hline
\textbf{Test Loss (PGD $L^\infty$)} & 9.73 & 5.98           \\ \hline
\end{tabular}
\label{tab:table_2}
\end{table}

\begin{table*}[htp!]
    \centering
    \caption{Networks performance after different attacking strategies. Both the networks were trained on PGD $L^2$ and PGD $L^\infty$ attacking strategies.}
    \label{tab:table_3}
    \begin{tabular}{|c|c|c|c|c|c|}
        \hline
        \textbf{Input image} & \textbf{Ground truth} & \textbf{UNet (PGD $L^2$)} & \textbf{UNet (PGD $L^\infty$)} & \textbf{LinkNet (PGD $L^2$)} & \textbf{LinkNet (PGD $L^\infty$)} \\
        \hline 
        \includegraphics[width=0.135\textwidth]{ 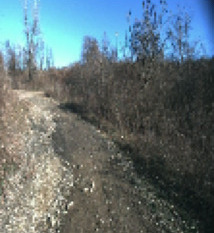} & \includegraphics[width=0.142\textwidth]{ 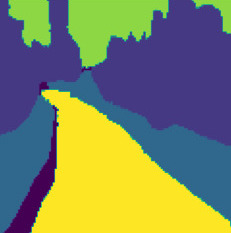} & \includegraphics[width=0.135\textwidth]{ 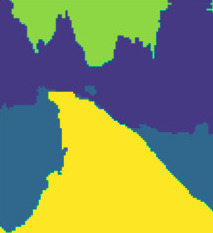} & \includegraphics[width=0.135\textwidth]{ 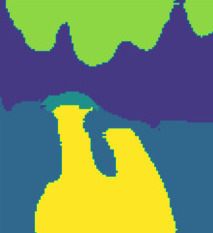} & \includegraphics[width=0.135\textwidth]{ 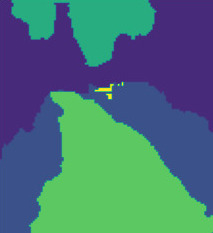} &
        \includegraphics[width=0.135\textwidth]{ 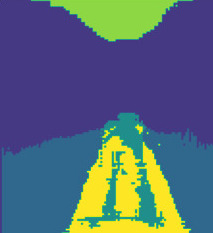} \\
        \hline    
        \includegraphics[width=0.135\textwidth]{ 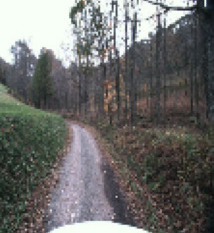} & \includegraphics[width=0.142\textwidth]{ 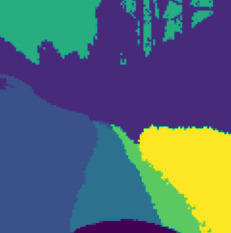} & \includegraphics[width=0.135\textwidth]{ 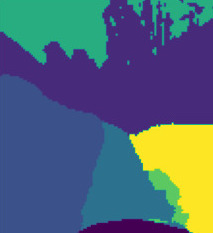} & \includegraphics[width=0.135\textwidth]{ 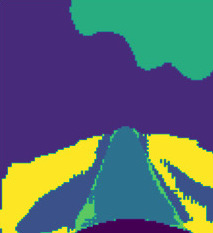} & \includegraphics[width=0.135\textwidth]{ 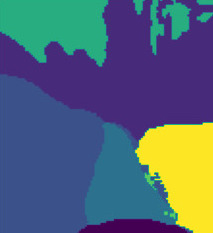} &
        \includegraphics[width=0.135\textwidth]{ 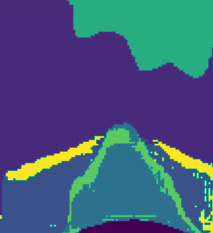} \\
        \hline 
    \end{tabular}
\end{table*}

\subsection{Adversarial training} \label{adversarial-training}
The goal of adversarial training is to obtain a classifier which will be robust to adversarial attacks and will serve as a basis for robustification procedure (see section \ref{sec:robust}). The adversarial training method (as suggested in \cite{ilyas2019adversarial}) is adapted to produce robust classifiers. Specifically, training against a PGD opponent is limited in the \textit{2-norm} starting from the original image. The gradient is calculated at each PGD step to make sure we move a specific amount in the \textit{2-norm} per step. At the end of this experiment, 4 robust models are obtained i.e. (2 models for \textit{PGD $L^2$} and 2 models for \textit{PGD $L^\infty$}). The quantitative evaluation of this training is further shown in table \ref{table_3}. 

\begin{table}[htp!]
\centering
\caption{Performance after 100 epochs of adversarial training with the attacking strategy.}
\begin{tabular}{|l|l|l|}
\hline
 & \textbf{UNet} & \textbf{LinkNet} \\ \hline
 \textbf{Train IoU (PGD $L^2$)}     & \textbf{0.68} & \textbf{0.68}           \\ \hline
\textbf{Train IoU (PGD $L^\infty$)}    & \textbf{0.50} & \textbf{0.45}        \\ \hline
\textbf{Train Loss (PGD $L^2$)}   & 0.33 & 0.37             \\ \hline
\textbf{Train Loss (PGD $L^\infty$)}  & 1.30 & 1.32          \\ \hline
\textbf{Val IoU (PGD $L^2$)}   & 0.68 & 0.68         \\ \hline
\textbf{Val IoU (PGD $L^\infty$)} & 0.52 & 0.46           \\ \hline
\textbf{Val Loss (PGD $L^2$)} & 0.28 & 0.32               \\ \hline
\textbf{Val Loss (PGD $L^\infty$)} & 1.36 & 1.35               \\ \hline
\end{tabular}
\label{table_3}
\end{table}

\subsubsection{\textbf{UNet trained with PGD $L^2$ and PGD $L^\infty$ attacks}}
The experiment details are as follows. UNet, with ResNet50 as backbone is trained for 100 epochs (with \textit{PGD $L^2$} and \textit{PGD $L^\infty$ attacks}) individually. \\When UNet (trained on \textit{PGD $L^2$}) is attacked independently with \textit{PGD $L^2$} and \textit{PGD $L^\infty$}, the results are presented in table~\ref{table_4}. When compared with the standard training (test dataset) evaluation (table \ref{tab:table_2}), an improvement is seen. The \textit{PGD $L^2$} loss for standard training was 0.59 whereas, it decreased to 0.37 during adversarial training. However, it is still evident that the model is susceptible to other attacks (e.g. performs badly with \textit{PGD $L^\infty$} attack). When compared with the results from standard training, it can be seen that, even though, quantitatively the model performs better, qualitatively, the model still lacks precision (see table \ref{tab:table_3}). The network performs comparatively poor when separating different surface types.  
\begin{table}[htp!]
\caption{Adversarially trained Unet and LinkNet models on two attack strategies are evaluated against both of these attack strategies. }
\resizebox{0.48\textwidth}{!}{%
\begin{tabular}{|l|l|l|l|l|}
\hline
 &
  \begin{tabular}[c]{@{}l@{}}\textbf{Test} \\ \textbf{IOU} (PGD $L^2$)\end{tabular} &
  \begin{tabular}[c]{@{}l@{}}\textbf{Test} \\ \textbf{IOU} (PGD $L^\infty$)\end{tabular} &
  \begin{tabular}[c]{@{}l@{}}\textbf{Test}\\ \textbf{Loss} (PGD $L^2$)\end{tabular} &
  \begin{tabular}[c]{@{}l@{}}\textbf{Test}\\ \textbf{Loss} (PGD $L^\infty$)\end{tabular} \\ \hline
 
\textbf{Unet trained on PGD $L^2$} &
  \textbf{0.71} &
  \textbf{0.60} &
  0.37 &
  6.71 \\ \hline
  
\textbf{UNet trained on PGD $L^\infty$} &
  \textbf{0.52} &
  \textbf{0.56} &
  1.27 &
  1.35 \\ \hline
  
  \textbf{Linknet trained on PGD $L^2$} &
  \textbf{0.61} &
  \textbf{0.70} &
  0.41 &
  5.18 \\ \hline
  
\textbf{Linknet trained on PGD $L^\infty$} &
  
  \textbf{0.55} &
  \textbf{0.58} &
  1.28 &
  1.31 \\ \hline
\end{tabular}%
}
\label{table_4}
\end{table}

When UNet (trained on \textit{PGD $L^\infty$}), is attacked similarly, the quantitative results are presented in table \ref{table_4} and the qualitative results can be seen from table \ref{tab:table_3}. It is visible that the loss (Train loss ($L^\infty$)) has increased whereas, the IoU has dropped significantly. On the contrary, it can be seen that the model trained on \textit{PGD $L^\infty$} also performs better with \textit{PGD $L^2$}. This is due to the fact that, \textit{PGD $L^2$} is a constrained form of \textit{PGD $L^\infty$} attacks and thus, the model shows more capability to learn constrained attacks automatically, if trained on broader attacks. 

\subsubsection{\textbf{LinkNet with PGD $L^2$ and PGD $L^\infty$ attacks}}
Similarly, the procedure is carried out with LinkNet and the results are presented in table \ref{table_4}. It is visible that the model performed better than the standard training (see table \ref{tab:table_2}).

When LinkNet (trained on \textit{PGD $L^2$}) is attacked with both \textit{PGD $L^2$} and \textit{PGD $L^\infty$}, the results are presented in table \ref{table_4}. It can be seen that the test loss has reduced significantly and the IoU has increased. At the same time, it also shows that the test loss for \textit{PGD $L^\infty$} is still significantly high, thereby confirming that the model does not learn the attacks if trained on one specific attack. \\
Similarly, LinkNet (trained on \textit{PGD $L^\infty$}) is attacked during test time and the results are presented in table \ref{table_4}. As visible, the test loss has decreased for both \textit{PGD $L^2$} and \textit{PGD $L^\infty$} thereby showing that training on broader attack families can help in acquiring immunity against smaller attacks of the same family.

\subsection{Dataset robustification} \label{sec:robust}
With the adversarially trained network as the classifier, in this experiment, the training dataset is robustified to construct a dataset that contains features relevant only to a given (standard/robust) model.\\
According to \cite{ilyas2019adversarial}, any given training example can be decomposed into robust and non-robust features. These subsets of features correspond to those that remain useful after adversarial perturbation and those that do not. The robust features are the ones that are truly important for determining the class of an object (e.g. eyes and ears of an animal). The non-robust features (according to humans) are more like background noise that can be easily changed without affecting the object's class. Generating a robust dataset begins by first creating a robust model through adversarial training as described in \ref{adversarial-training}. The robustified model includes all but the output layer of the adversarially trained model. The representation for a data point is the output of this penultimate layer. For each data point in the original dataset, a random starting input is iteratively updated to create a representation that is as similar to the original input as possible. The perturbation process aims to minimize the $L^2$ distance between the label and the sample's representation. Because the model itself is robust, the adjustments to the random input only reflect the robust features of the original data point.

\begin{algorithm}
\small
\caption{Pseudo-code for creating robust dataset (D). \cite{ilyas2019adversarial}}
\begin{algorithmic}[1]
\State $C_R \gets \text{AdversarialTraining}(D)$ 
\State $g_R \gets \text{mapping learned by $C_R$ from input to the representation layer}$ 
\State $D_R \gets \{\}$ 
\For {$(x, y) \in D$} 
    \State $x' \sim D$
    \State $x_R \gets \text{arg min}_{z \in [0,1]^d} \lVert g_R(z) - g_R(x) \rVert_2$ 
    \State $D_R \gets D_R \cup \{(x_R, y)\}$ 
\EndFor
\State \textbf{Return} $D_R$ 
\end{algorithmic}
\end{algorithm}

Once this robustified model has been created, the next step is to make a dataset containing only the features relevant to the specific model. To achieve this, the input $x_r$ is initialized as a different randomly chosen sample from the training set. Normalized gradient descent is then performed on $x_r$ while fixing the $L^2$-norm of the gradient at each step to a constant value. This helps to ensure that the perturbations made to $x_r$ are not too large and do not cause significant changes to the image. At each step of the process, the input $x_r$ is clipped so that it falls within the valid image range of [0, 1] thereby maintaining the integrity of the images and ensuring that it remains recognizable as an object within the given dataset.\\
To perform this step, several iterations (100) based on batch size are performed. For this, the configurations mentioned in table \ref{tab:table_2} are taken. In each iteration, a batch of training data is passed through the robust model to obtain the goal representation. The goal representation is then attacked using the PGD attack, resulting in a delta on the input image batch. The delta is then concatenated with the original image and saved to create a robust dataset that includes both the training images and their corresponding labels. This is saved and evaluated on the standard models against the PGD attacks.

At the end of the experiment, we obtain robustified dataset against \textit{PGD $L^2$ and PGD $L^\infty$} attack. With 4 adversially trained models , 4 robustified datasets are obtained i.e.
\begin{itemize}
    \item Adversarially trained UNet + \textit{PGD $L^2$} robustified dataset
    \item Adversarially trained UNet + \textit{PGD $L^\infty$} robustified dataset
    \item Adversarially trained LinkNet + \textit{PGD $L^2$} robustified dataset
    \item Adversarially trained LinkNet + \textit{PGD $L^\infty$} robustified dataset
\end{itemize}

\begin{table*}[htp!]
    \centering
    \caption{Networks performance on the robustified dataset. \textit{R} represents robustified dataset. Networks struggle to capture the details of an ambiguous environment.}
    \label{tab:table_7}
    \begin{tabular}{|c|c|c|c|c|c|}
        \hline
        \textbf{Input image} & \textbf{Ground truth} & \textbf{UNet (RPGD $L^2$)} & \textbf{UNet (RPGD $L^\infty$)} & \textbf{LinkNet (RPGD $L^2$)} & \textbf{LinkNet (RPGD $L^\infty$)} \\
        \hline 
        \includegraphics[width=0.135\textwidth]{ 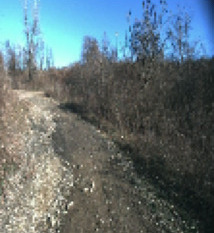} & \includegraphics[width=0.142\textwidth]{ 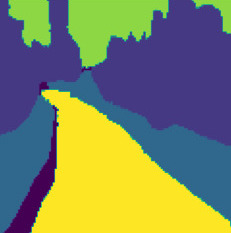} & \includegraphics[width=0.135\textwidth]{ 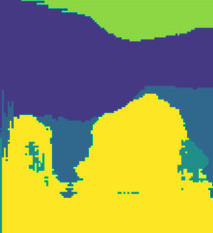} & \includegraphics[width=0.135\textwidth]{ 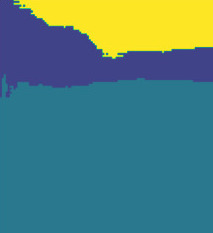} & \includegraphics[width=0.135\textwidth]{ 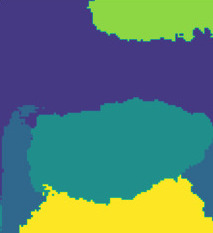} &
        \includegraphics[width=0.135\textwidth]{ 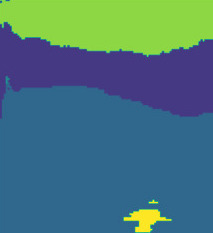} \\
        \hline    
        \includegraphics[width=0.135\textwidth]{ 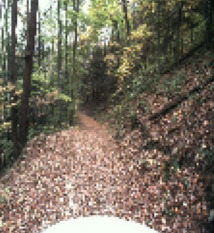} & \includegraphics[width=0.142\textwidth]{ 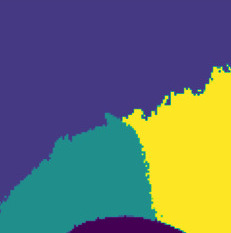} & \includegraphics[width=0.135\textwidth]{ 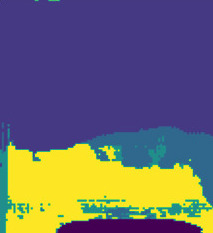} & \includegraphics[width=0.135\textwidth]{ 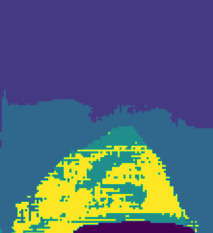} & \includegraphics[width=0.135\textwidth]{ 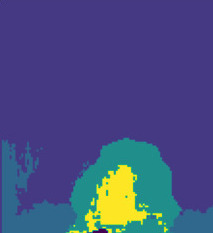} &
        \includegraphics[width=0.135\textwidth]{ 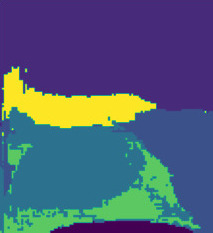} \\
        \hline 
         \includegraphics[width=0.135\textwidth]{ images/unetL2/tile-6.jpg} & \includegraphics[width=0.142\textwidth]{ images/unetL2/tile-7.jpg} & \includegraphics[width=0.135\textwidth]{ 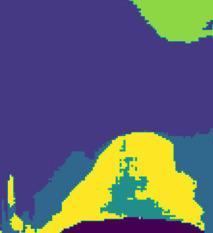} & \includegraphics[width=0.135\textwidth]{ 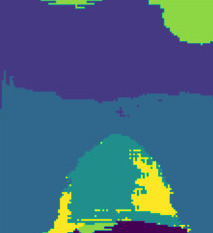} & \includegraphics[width=0.135\textwidth]{ 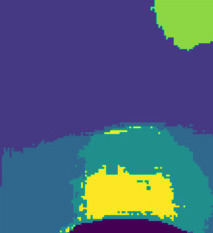} &
        \includegraphics[width=0.135\textwidth]{ 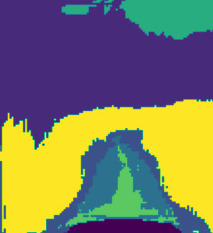} \\
        \hline
         \includegraphics[width=0.135\textwidth]{ 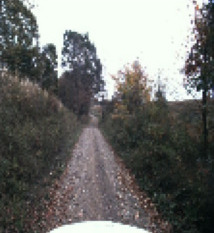} & \includegraphics[width=0.142\textwidth]{ 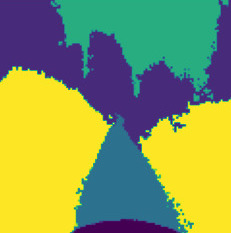} & \includegraphics[width=0.135\textwidth]{ 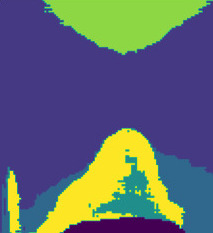} & \includegraphics[width=0.135\textwidth]{ 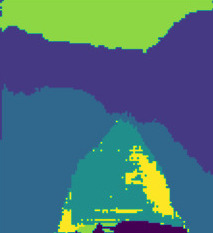} & \includegraphics[width=0.135\textwidth]{ 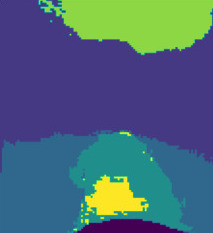} &
        \includegraphics[width=0.135\textwidth]{ 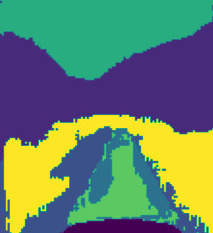} \\
        \hline
    \end{tabular}
\end{table*}

\begin{table*}[htp!]
    \centering
    \caption{16 Feature activations of the second last convolution layer depicting the network's attention during standard training, adversarial training and standard training on robustified dataset. Robustified dataset being depicted by \textit{R}.}
    \label{tab:table_8}
    \begin{tabular}{|c|c|c|c|c|}
        \hline
        \textbf{UNet-standard training} & \textbf{UNet-PGD $L^2$} & \textbf{UNet-PGD $L^\infty$} & \textbf{UNet-RPGD $L^2$} & \textbf{UNet-RPGD $L^\infty$} \\
        \hline 
        \includegraphics[width=0.17\textwidth]{ 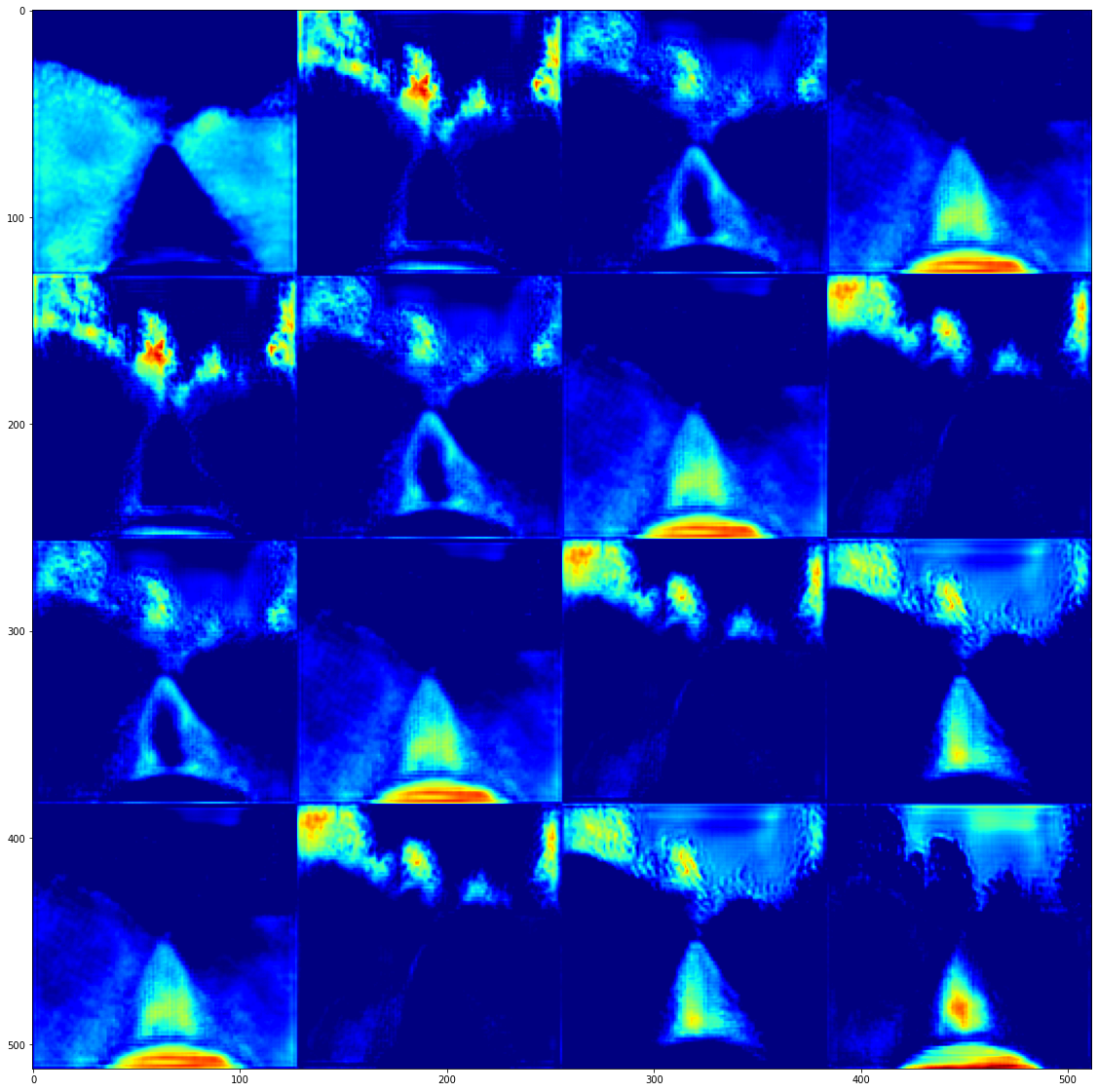} & \includegraphics[width=0.17\textwidth]{ 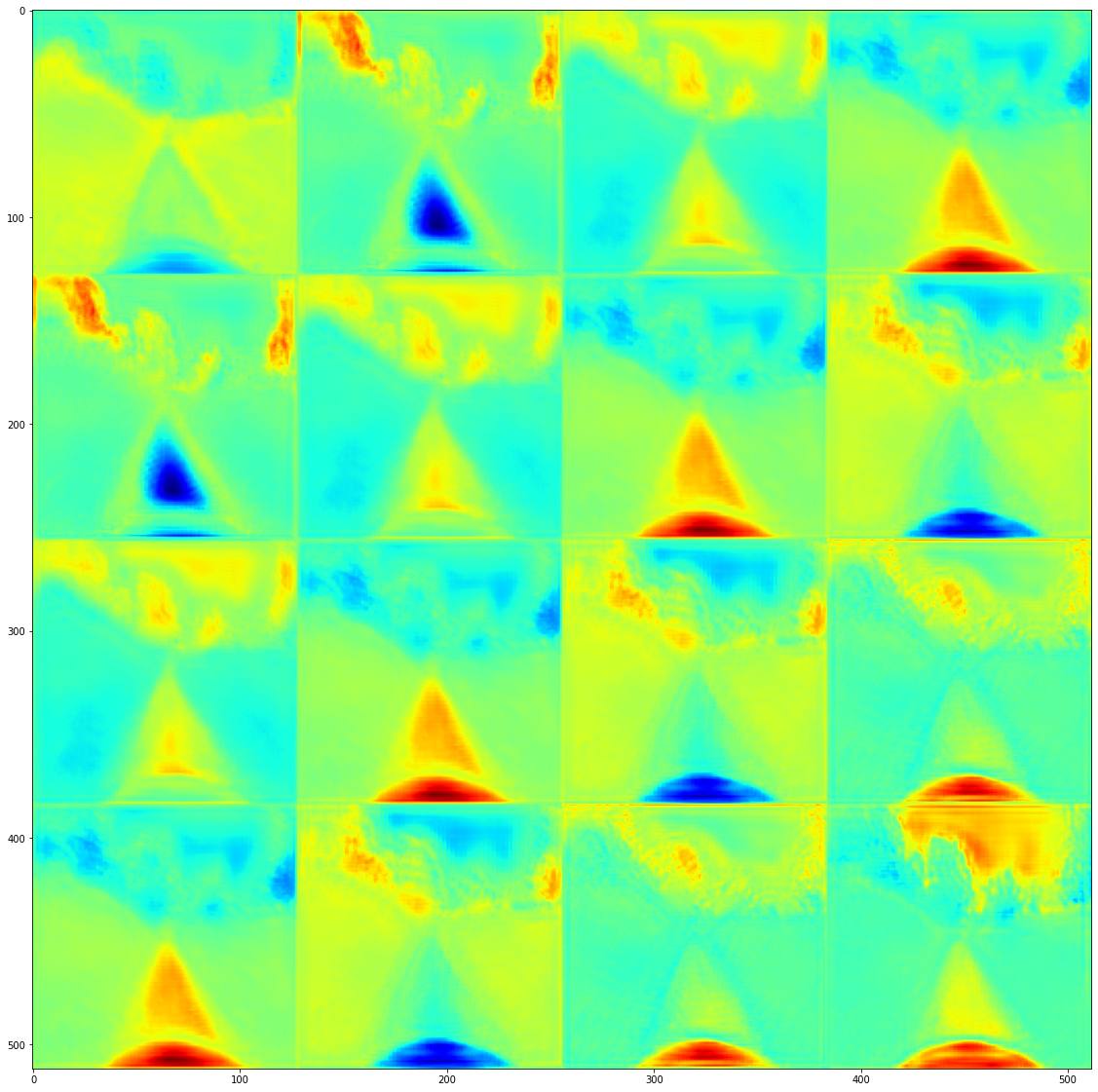} & \includegraphics[width=0.17\textwidth]{ 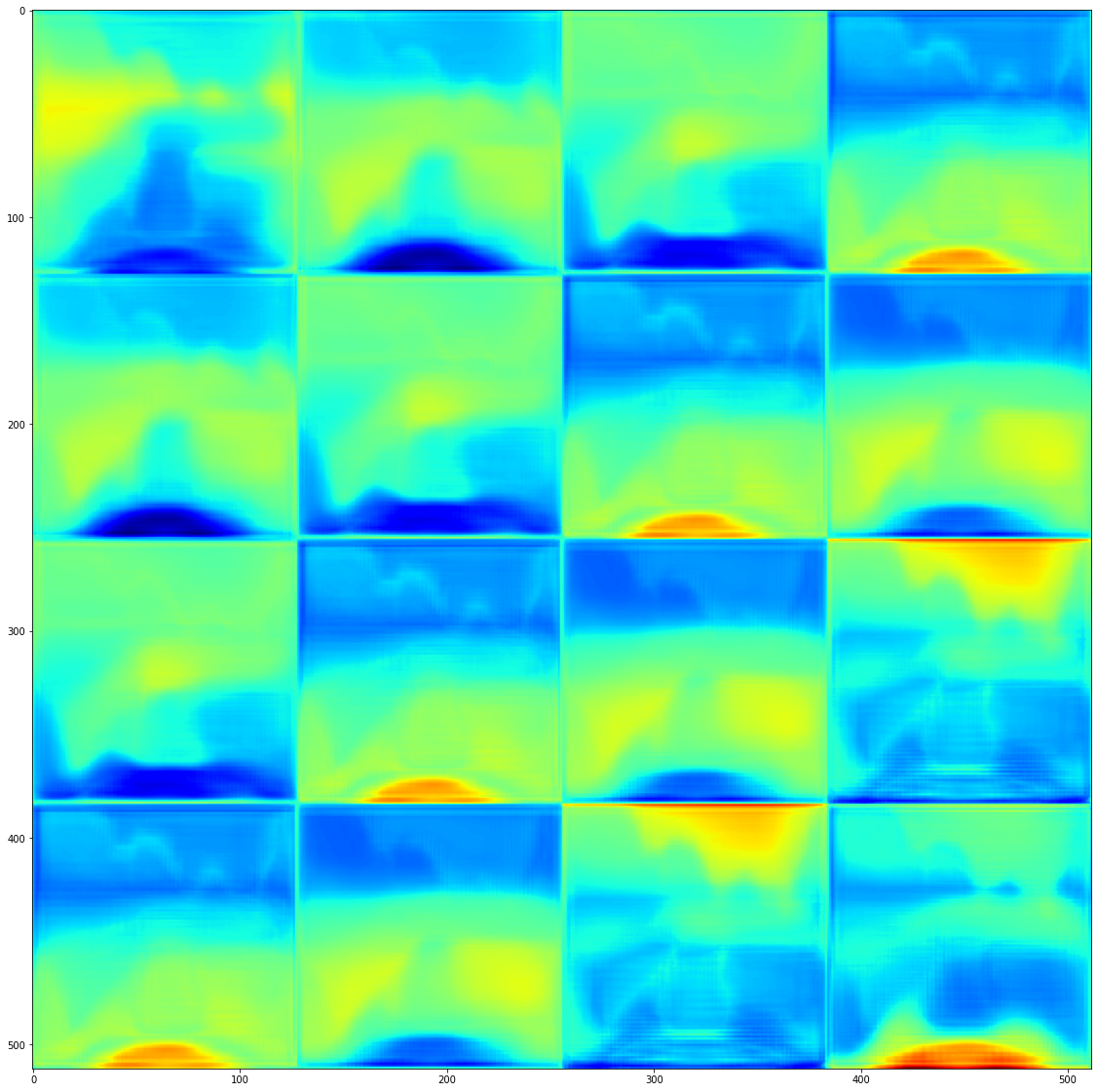} & \includegraphics[width=0.17\textwidth]{ 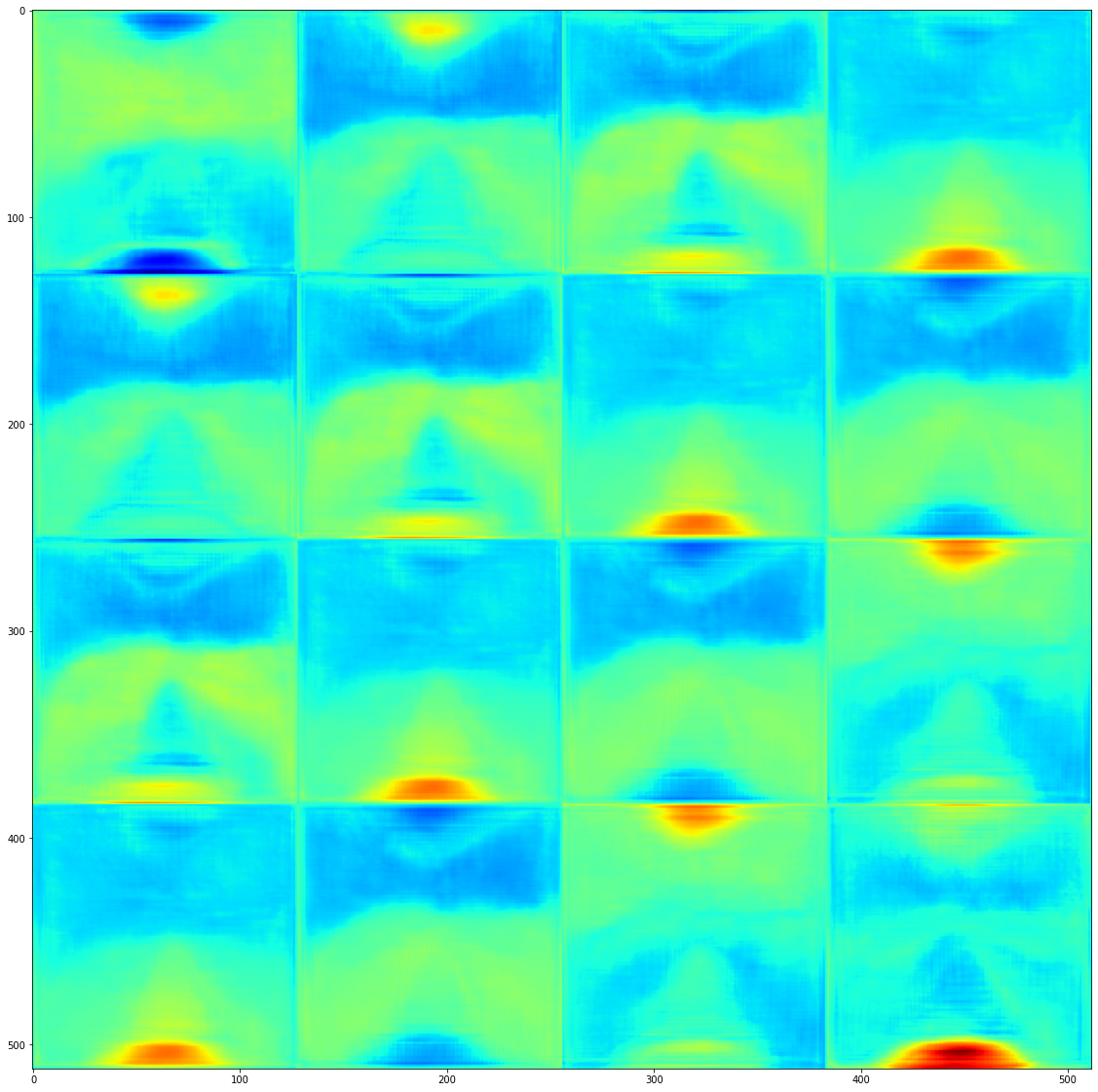} & \includegraphics[width=0.17\textwidth]{ 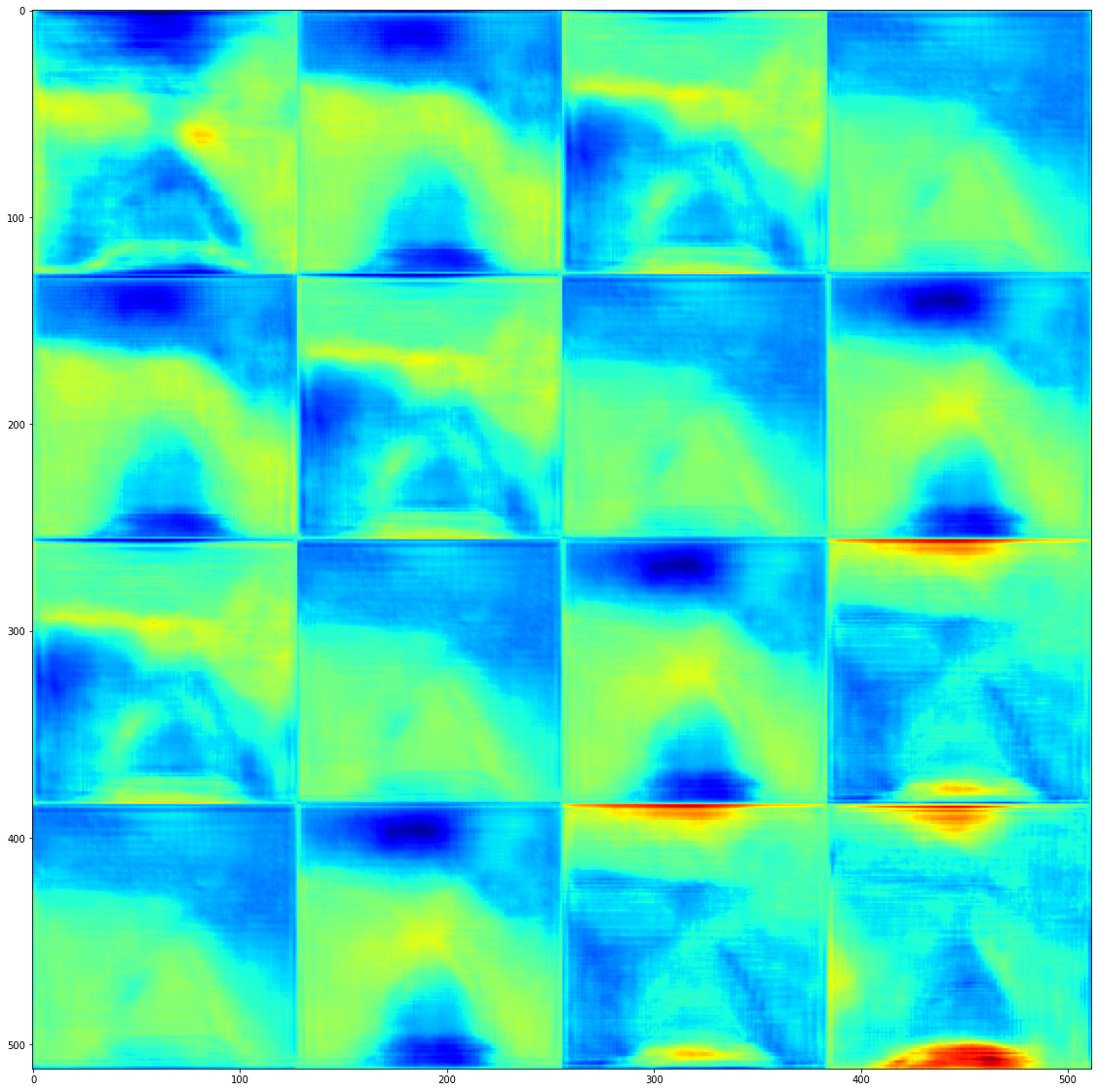}  \\
        \hline 
    \end{tabular}
\end{table*}

\subsection{Standard training on Robustified dataset}
In the final experiment, the robustified dataset obtained from the adversarially trained classifiers is taken for standard training with a total of 4 networks (obtained from adversarial training). No attack (of any kind) is employed during training as well as test time. The models are trained for 100 epochs on the robust dataset and validated on the original validation dataset.  
\subsubsection{\textbf{UNet}}
Both the UNet robustified datasets are selected for standard training. As seen from table \ref{tab:unet-robust-train}, the results are comparable to the standard training results (from experiment~\ref{standard-test}) thereby depicting that the model learns the features from the robustified training set. However, the losses and IoU values for the original validation dataset are not efficient (for both the robustified datasets). Qualitatively (table \ref{tab:table_7} (column c and d), the model fails to converge and predict proper segmentation, thereby depicting that even though, the quantitative results are promising, the actual qualitative results are misleading.

\begin{table}[!htp]
\caption{Standard training of robustified UNet and LinkNet model. The first \& third row shows results on the robustified dataset with PGD $L^2$ and the second \& fourth row shows results on the robustified dataset with PGD $L^\infty$ at train time.}
\resizebox{0.47\textwidth}{!}{%
\begin{tabular}{|l|l|l|l|l|}
\hline
    &
 \begin{tabular}[c]{@{}l@{}}\textbf{Train IOU}\end{tabular} &
  \begin{tabular}[c]{@{}l@{}}\textbf{Val IOU}\end{tabular} &
  \begin{tabular}[c]{@{}l@{}}\textbf{Train Loss} \end{tabular} &
  \begin{tabular}[c]{@{}l@{}}\textbf{Val Loss}\end{tabular}  \\ \hline
  
\textbf{UNet (PGD $L^2$)} &
  \textbf{0.68} &
  0.58 &
  0.22 &
  2.44 \\ \hline
  
\textbf{UNet (PGD $L^\infty$} &
  \textbf{0.72} &
  0.57 &
  0.26 &
  2.40 \\ \hline
  
  \textbf{LinkNet (PGD $L^2$)}     & \textbf{0.70}      & 0.60 & 0.26       & 2.02         \\ \hline
\textbf{LinkNet (PGD $L^\infty$)} & \textbf{0.71}      & 0.58   & 0.29       & 2.06       \\ \hline
\end{tabular}%
}
\label{tab:unet-robust-train}
\end{table}

\subsubsection{\textbf{LinkNet}}
Similar procedure for LinkNet is performed, the results of which can be seen from table \ref{tab:table_7} (column e and f). The model shows similar performance as UNet wherein, the qualitative and quantitative results are misleading.

\section{Findings}
The work explored whether the dataset has any effect in countering the adversarial effect on semantic segmentation tasks and whether it can be changed (based on attacks) and still maintain the same performance w.r.t state-of-the-art metrics. One conclusion, which is evident, is the direct unreliable relation between qualitative and quantitative performance. The study done by \cite{ilyas2019adversarial} depicts that the robustified model performs better, however, the qualitative results were not talked about since it focused on classification rather than segmentation. From off-road autonomous driving point-of-view, some attributes towards the performance of our chosen models could be:
\begin{itemize}
    \item \textbf{Presence of Multi-class in input images} - The previous study by \cite{ilyas2019adversarial}, in this regard, concentrated on binary classification. However, in our study, the presence of ambiguous (non-rich features) multi-classes of the input space can present fundamental barriers to classifier's robustness. A classifier cannot be resistant against tiny perturbations since, at the highest level, for certain data distributions, any decision boundary will be near a significant portion of inputs. Further details of these can be visualized by (see table \ref{tab:table_8})
    where is clearly visible that due to the presence of many ambiguous classes, the network fails to be persistent.
    \item \textbf{Insufficient data} - Another major problem was the unavailability of a huge, distinct dataset with higher variability. Authors of \cite{schmidt2018adversarially} present that for appropriate learning, a good robust classifier requires (\textit{O$\sqrt{d}$}) samples ($d$ being the dimentionality of the data). In this paradigm, adversarial examples appear as a result of insufficient knowledge of the real data distribution. In particular, since training models robustly reduce the effective amount of information in the training data (as non-robust features are discarded), more samples should be required to generalize robustly.
\end{itemize}

\section{What does it mean  for the Unimog U5023?}

To accurately analyse the developed solutions, Unimog U5023 has been used as a demonstrator vehicle. The vehicle has the capability of accommodating adverse off-road conditions and can therefore be also used as a method of methodological transfer (to other smaller robots). The vehicle has additional D.O.F (Degrees of freedom), a controllable gearbox, differential locks, or adaptable tire pressure. For encompassing sudden terrains, the frame has the capability to bend diagonally up to 60 cm while the axle can be articulated up to 30 degrees. 

\begin{figure}[htp!]
    \centering
    \includegraphics[width=0.47\textwidth]{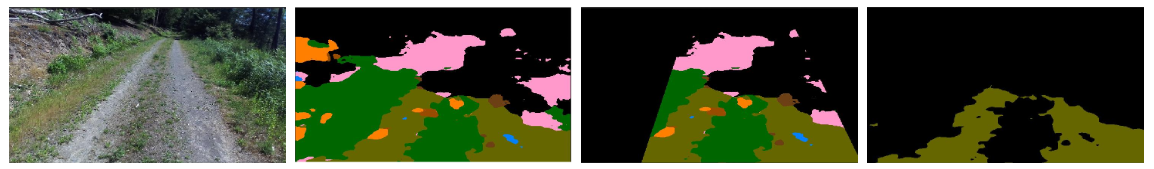}
    \caption{The results for surface type detection in \cite{wolf_pankaj_2023}. Image represents input image, segmentation result, Region of Interest and the dominant class taken. }
    \label{fig:traction}
\end{figure}

Previous works done \cite{wolf_pankaj_2023}, \cite{wolf_patrick}, \cite{9636541} towards safety evaluation have focused on robust environment perception for safe navigation in off-road unstructured environments. In \cite{wolf_pankaj_2023}, a methodology was developed for traction optimization for Unimog based on OSM maps and semantic segmentation (shown in figure \ref{fig:traction}). With the help of a neural network, the slip coefficient of the segmented pathway is calculated which is compared with the OSM tag for the corresponding surface. With the help of precise segmentation, the Unimog can be traversed through an unstructured environment with the correct traction control ultimately providing an extra layer of safety.  

The Unimog is equipped to perform various tasks in an off-road environment such as forest path making with a shredder, monotonous tasks in a landfill and performing rescue operations after natural calamities. The most important factor when performing these operations is the appropriate understanding of the surface type, since a small slip factor (traction) change might result in a sudden speed shift. The pre-existing surface type detection module [24] has the capabilities of separating surfaces according to its surface properties. However, to make this more robust, a subset of this research will be used to focus on neural network training only on robust features to make better surface type detection. Since off-road environment is ambiguous in nature (i.e. poor distinction between foreground and background) training on less no. of classes (e.g. 2 or 3 types of surfaces) will force the network to understand the salient features within the surface paradigm thereby providing improved robustness during autonomous navigation tasks and subsequently improving the safety of the Unimog.

\section{conclusions}
In conclusion, the work explored the role of datasets for achieving the segmentation robustness in off-road environments. Two SOTA neural networks along-with a combined dataset was utilised to see the effects when these are presented by adversarial attacks. Even though, the work presented a method to generalize the classifier's robustness, the analysis is still at a nascent stage as seen from the unreliable results between qualitative and quantitative results. Even while these evaluation criteria exceeded industry standards
and performed well when compared to other studies in related fields, it was discovered
that they do not accurately reflect the complexity of the off-road environment. To mimic human visual system, classifiers require large-norm perturbations to be fooled. Reaching such levels require careful selection of prior distribution, smoothness and dimension attributes. Further, intelligent redundancy and ensembles of model-agnostic defences should be analysed in fending off-adversarial attacks.


%






\bibliographystyle{ACM-Reference-Format} 
\bibliography{main}


\end{document}